\documentclass[11pt]{article}
\usepackage{coling2020}
\usepackage{times}
\usepackage{url}
\usepackage{latexsym}

\usepackage{multirow}
\usepackage{multicol}
\usepackage{array}
\usepackage{booktabs}
\usepackage{hyperref}

\usepackage[colorinlistoftodos]{todonotes}
\usepackage{amsmath}
\colingfinalcopy 

\title{MTLB-STRUCT @PARSEME 2020: Capturing Unseen Multiword Expressions Using Multi-task Learning and Pre-trained Masked Language Models
}

\author{Shiva Taslimipoor \\
  ALTA Institute \\
  University of Cambridge, UK \\
  {\tt st797@cl.cam.ac.uk} \\\And
  Sara Bahaadini \\
  Microsoft  \\
  Sunnyvale, CA, USA \\
  {\tt sabahaa@microsoft.com} \\\And
  Ekaterina Kochmar \\
  ALTA Institute \\
  University of Cambridge, UK \\
  {\tt ek358@cl.cam.ac.uk} }

\date{}

\begin{document}
\maketitle
\begin{abstract}
This paper describes a semi-supervised system that jointly learns verbal multiword expressions (VMWEs) and dependency parse trees as an auxiliary task. The model benefits from pre-trained multilingual BERT. 
BERT hidden layers are shared among the two tasks and we introduce an additional linear layer to retrieve VMWE tags. The dependency parse tree prediction is modelled by a linear layer and a bilinear one plus a tree CRF on top of BERT. The system has participated in the open track of the PARSEME shared task 2020 and ranked first in terms of F1-score in identifying unseen VMWEs as well as VMWEs in general, averaged across all $14$ languages.  
  
\end{abstract}

\section{Introduction}
\label{intro}

In addition to other challenges in multiword expression (MWE) processing that were addressed in previous work, such as non-compositionality \cite{salehi2014}, discontinuity \cite{rohanian2019bridging,waszczuk2018}, and syntactic variability \cite{pasquer2018}, The PARSEME shared task edition 1.2\footnote{\url{http://hdl.handle.net/11234/1-3367}} has focused on another prominent challenge in detecting MWEs, namely detection of unseen MWEs. The problem with unseen data is common for many NLP tasks. While rule-based and unsupervised ML approaches are less affected by unseen data, supervised ML techniques are often found to be prone to overfitting. In this respect, the introduction of language modelling objectives to be added to different NLP tasks and their effect on generalisation have shown promising results~\cite{rei2017-semi}. Further improvements brought by pre-trained language models made them a popular approach to a multitude of NLP tasks~\cite{devlin2018bert}. One particular advantage of such models is that they facilitate generalisation beyond task-specific annotations \cite{pires2019multilingual}.

MWEs are inherent in all natural languages and distinguishable for their syntactic and semantic idiosyncracies \cite{baldwin2010,fazly2009unsupervised}. Since language models are good at capturing syntactic and semantic features, we believe they are a suitable approach for modelling MWEs.   
In particular, our system relies on BERT pre-trained language models \cite{devlin2018bert}. 
Additionally, we render the system semi-supervised by means of multi-task learning. The most promising feature to be jointly learned with MWEs is dependency parse information \cite{constant2016transition}. Accordingly, we fine-tune BERT for two different objectives: MWE detection and dependency parsing. MWE learning is done via token classification using a linear layer on top of BERT, and dependency parse trees are learned using dependency tree CRF network \cite{rush-2020-torch}. 
Our experiments confirm that this joint learning architecture is effective for capturing MWEs in most languages represented in the shared task.~\footnote{The code for the system and configuration files for different languages are available at \url{https://github.com/shivaat/MTLB-STRUCT/}}

\section{Related Work}

In earlier systems, MWEs were extracted using pre-defined patterns or statistical measures that either indicated associations among MWE components or (non-)compositionality of the expressions with regard to the components \cite{ramisch2010mwetoolkit}. For example, \newcite{cordeiro2016} employed such a system for identifying MWEs. 
While these models can be effective for some frequent MWEs, their main disadvantage is that they capture MWE types (as opposed to tokens) and they are unable to take context into account in running texts. 

The use of supervised machine learning was facilitated by the availability of resources tagged for MWEs \cite{schneider2014,savary2017parseme,ramisch2018edition}. \newcite{al2017atilf} proposed a transition-based system based on an arc-standard dependency parser \cite{nivre2004} which ranked first in the first edition of PARSEME shared task on automatic identification of verbal MWEs (VMWEs)~\cite{savary2017parseme}. \newcite{taslimipoor2018shoma} proposed a CNN-LSTM system which exploited fastText word representations and ranked first in the open track of the PARSEME shared task edition 1.1~\cite{ramisch2018edition}. 
Previous systems such as TRAVERSAL \cite{waszczuk2018} (ranked first in the closed track of the PARSEME shared task edition 1.1), and CRF-Seq/Dep \cite{moreau2018crf} employed tree CRF using dependency parse features in non-deep learning settings. They showed strengths of this approach particularly in the case of discontinuous VMWEs. In SHOMA \cite{taslimipoor2018shoma},  using a linear-chain CRF layer on top of the CNN-biLSTM model did not result in improvements.
In this work, we use tree CRF, implemented as part of the Torch-Struct library \cite{rush-2020-torch}, to model dependency trees, and we show that when it is jointly trained with a transformer-based MWE detection system, it improves MWE prediction for a number of languages.

Recently, \newcite{savary2019without} proposed that learning MWE lexicons in an unsupervised setting is an important step that can be used in combination with a supervised model, especially when the latter is trained on a small amount of data. While we do not specifically learn MWE lexicons from external unannotated data, we believe that state-of-the-art pre-trained language representation models can capture crucial information about MWEs similar to other NLP phenomena \cite{peters2017semi}. For instance, \newcite{peters2017semi} showed how a semi-supervised system may benefit from pre-trained language model-based embeddings for named entity recognition (NER) and chunking. 
The joint learning of MWEs and dependency parsing has been proved effective in \newcite{constant2016transition}. They proposed an arc-standard transition-based system which draws on a new representation that has two linguistic layers (a syntactic dependency tree and a forest of MWEs) sharing lexical nodes.  The closest to our work is \newcite{taslimipoor2019cross} where they have trained a multi-task neural network which jointly learns VMWEs and dependency parsing on a small English dataset and uses ELMo pre-trained embeddings. %
Our work here is different in that we fine-tune the BERT architecture and we use a tree CRF for dependency parsing. 

\section{System Description}
\label{sec:description}
We use pre-trained BERT for language representation \cite{devlin2018bert} as the basis for our neural network.  
The BERT architecture is based on standard transformers involving self-attention layers of encoders and decoders.\footnote{There are 12 layers (transformer blocks) following the implementation of \url{http://nlp.seas.harvard.edu/2018/04/03/attention.html}, with the hidden dimension size of 768 and 12 attention heads.} What makes it different from other transformer-based pre-trained language representation models is its capability in encoding the representation in a bidirectional way through a masked language model schema. The reason that we choose BERT among other pre-trained models is the availability of multi-lingual pre-trained BERT.\footnote{\url{https://huggingface.co/bert-base-multilingual-cased}} 

\begin{figure}
\centering
\includegraphics[width=10.cm,height=3.87cm]{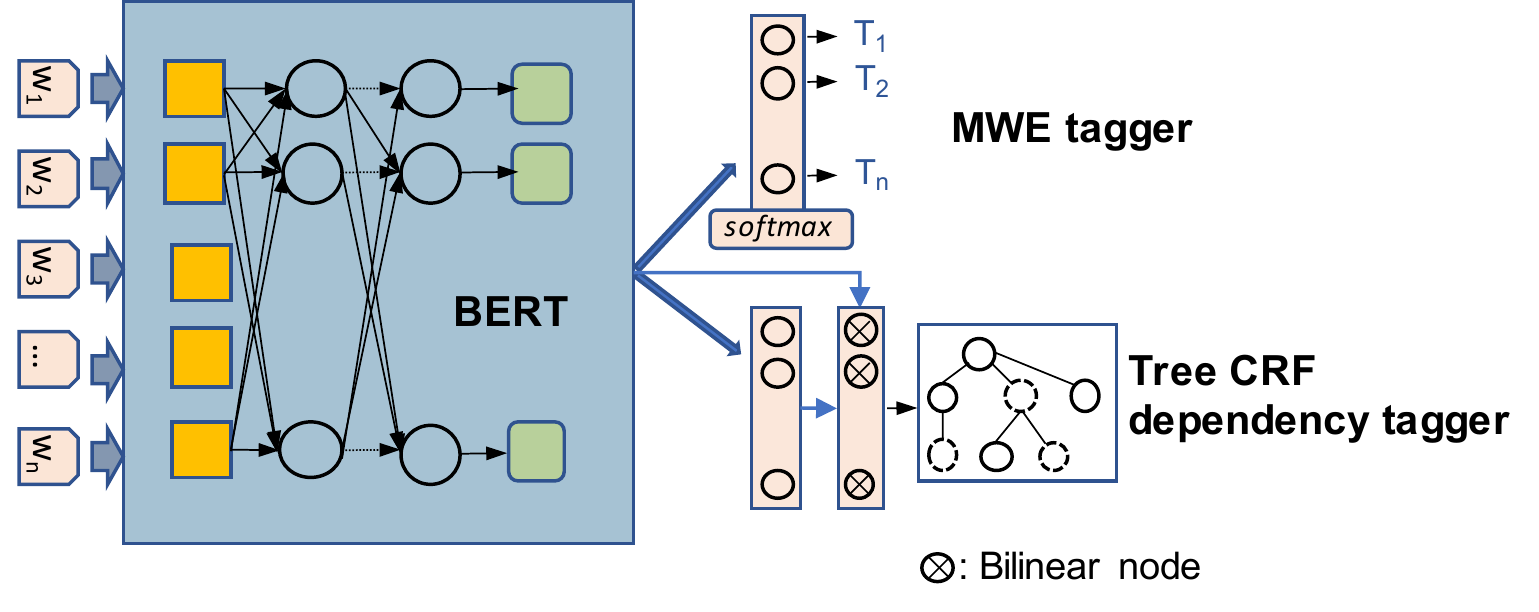}
\vspace*{-\baselineskip}
\caption{The overall architecture of the multi-task learning model with two branches on top of BERT. One is a linear classifier layer for MWE tagging and the other consists of a linear layer, a bilinear layer and a tree CRF dependency tagger.}
\label{fig:model}
\vspace*{-3mm} 
\end{figure}

Our model is set up to learn MWEs and dependency trees simultaneously. BERT weights are shared among the two tasks. A fully connected layer that performs sequence tagging is added as the final layer for MWE objective. Parallel to that, linear layers and a dependency CRF module are introduced to perform structured prediction for dependency trees.~\footnote{In this work, we only focus on dependency arcs (tree structures) and we do not model dependency relation labels.} The whole model is trained in an end-to-end manner. 
Figure \ref{fig:model} depicts the overall architecture of the system. 


We use Torch-Struct \cite{rush-2020-torch} for dependency parsing where Tree CRF is implemented as a distribution object. We first apply a linear followed by a bilinear layer on BERT's output to obtain the adjacency matrix structure of the dependency tree. 
The outputs from these layers are considered as log-potentials ($l$) for the CRF distribution. The distribution takes in log-potentials and converts them into probabilities $CRF(z;l)$ of a specific tree $z$. We query the distribution to predict over the set of trees using $arg max_z CRF(z;l)$. The cost for updating the tree is based on the difference between the tree probability and the gold standard dependency arcs.

The MWE classification layer is optimised by cross-entropy between the ground truth MWE tags and the predicted ones, while the cost for CRF is estimated using log probabilities over the tree structures. 
Note that log probabilities ($logprobs$) for CRF are large negative values which should be maximised, so we multiply them by $-1$ to get the dependency loss values compatible with MWE ones: $Loss_{dep} = - {logprobs}$.  
The overall loss function to be optimised by ADAM optimiser is a linear combination of the two losses, $Loss_{mwe}$ and $Loss_{dep}$ which are the losses for multi-word expression and dependency parse tree, respectively, with $\alpha$ being a constant value which is empirically set to $  0.001 \leq \alpha \leq 0.01$.
\vspace*{-3mm} 

 \begin{equation}
 {Loss} = Loss_{mwe} + \alpha * Loss_{dep}
 \label{eq:loss}
 \end{equation}


\begin{table}[t]
\centering
\small

\begin{tabular}{|l l | c c c | c c c | c |}
\toprule
\multirow{2}{*}{} & \multirow{2}{*}{System} & \multicolumn{3}{c}{Global MWE-based} & \multicolumn{3}{c}{Unseen MWE-based} & Variant  \\
  & & P & R & F1 & P & R & F1 & F1 \\
\midrule
\multirow{2}{*}{DE} & single-task (bert German cased) & 79.45 & 75.28 &  77.31 & 53.00 & 53.00 & \textbf{53.00} & 90.32 \\ 
& multi-task (bert German cased) 
& 74.81 & 75.66 & 75.23 & 46.61 & 55.00 & 50.46 & 90.91 \\
& single-task (bert multilingual) & 73.06 & 74.16 & 73.61 & 45.45 & 55.00 & 49.77 & 86.96\\
\midrule
\multirow{2}{*}{EL} & single-task & 70.28 & 72.06 & 71.16 & 37.01 & 47.00 & 41.41 & 81.98  \\
& multi-task 
& 72.38 & 72.38 & 72.38 & 40.68 & 48.00 & \textbf{44.04} & 82.30 \\ 
\midrule
\multirow{2}{*}{EU} & single-task & 76.82 & 78.54 & 77.68 & 28.10 & 43.00 & 33.99 & 81.20   \\
& multi-task 
& 79.04 & 80.22 & 79.63 & 29.08 & 41.00 & \textbf{34.02} & 83.78 \\
\midrule
\multirow{2}{*}{FR} & single-task & 83.17 & 79.06 & 81.06 & 48.57 & 50.50 & \textbf{49.51} & 87.05 \\
& multi-task 
& 81.53 & 80.00 & 80.76 & 44.92 & 52.48 & 48.40 & 87.46 \\
& *single-task (camembert) & 79.61 & 85.41 & 82.41 & 45.32 & 62.38 & 52.50* & 91.16 \\
\midrule
\multirow{2}{*}{GA} & single-task & 26.15 & 13.49 & 17.80 & 16.07 & 9.00 & 11.54  & 32.00  \\
& multi-task 
& 25.00 & 14.29 & 18.18 & 18.18 & 12.00 & \textbf{14.46} & 25.00 \\
\midrule
\multirow{2}{*}{HE} & single-task & 52.76 & 40.36 & 45.73 & 23.94 & 16.67 & 19.65 & 73.91 \\
& multi-task 
& 57.14 & 38.55 & 46.04 & 31.15 & 18.63 & \textbf{23.31} & 58.54 \\
\midrule
\multirow{2}{*}{HI} & single-task & 71.78 & 62.90 & 67.05 & 50.55 & 46.00 & \textbf{48.17} & 83.12  \\
& multi-task 
& 64.09 & 62.37 & 63.22 & 39.62 & 42.00 & 30.78 & 87.50  \\
\midrule
\multirow{2}{*}{IT} & single-task & 70.53 & 62.04 & 66.01 & 32.35 & 32.67 & \textbf{32.51} & 76.02  \\
& multi-task 
& 71.84 & 61.42 & 66.22 & 29.90 & 28.71 & 29.29 & 78.90 \\
\midrule
\multirow{2}{*}{PL} & single-task & 83.22 & 81.72 & 82.46 & 42.24 & 49.00 & \textbf{45.37} & 91.51  \\
& multi-task 
& 83.92 & 80.14 & 81.99 & 42.59 & 46.00 & 44.23 & 90.30  \\
\midrule
\multirow{2}{*}{PT} & single-task & 78.82 & 74.06 & 76.36 & 33.64 & 36.00 & 34.78 & 87.85  \\
& multi-task 
& 80.11 & 73.05 & 76.42 & 40.40 & 41.00 & \textbf{40.59} & 84.08  \\
\midrule
\multirow{2}{*}{RO} & single-task & 91.41 & 85.82 & 88.52  & 39.13 & 36.00 & \textbf{37.50} & 85.39  \\
& multi-task 
& 91.07 & 86.06 & 88.50 & 39.53 & 34.00 & 36.56 & 85.29 \\
\midrule
\multirow{2}{*}{SV} & single-task & 68.99 & 65.93 & 67.42 & 39.83 & 47.00 & 43.12 & 77.92  \\
& multi-task 
& 70.37 & 70.37 & 70.37 & 41.13 & 51.00 & \textbf{45.54} & 81.53\\
\midrule
\multirow{2}{*}{TR} & single-task & 62.59 & 69.75 & 65.98 & 37.41 & 52.00 & 43.51 & 68.23  \\
& multi-task 
& 66.06 & 69.48 & 67.73 & 43.31 & 55.00 & \textbf{48.46} & 66.43 \\
& *multi-task 
(+ extra data) & 67.89 & 70.84 & 69.33 & 45.08 & 55.00 & 49.55* & 70.79 \\
\midrule
\multirow{2}{*}{ZH} & single-task & 72.39 & 73.21 & 72.80 & 59.13 & 60.18 & 59.65 & 71.43 \\
& multi-task 
& 72.45 & 72.45 & 72.45 & 60.36 & 59.29 & \textbf{59.82} & 71.43 \\
& *single-task (bert Chinese cased) & 73.14 & 78.11 & 75.55 & 61.07 & 70.80 & 65.57* & 71.43 \\
& *multi-task (bert Chinese cased) & 70.92 & 75.47 & 73.3 & 50.68 & 65.49 & 62.45 & 80.00 \\
\bottomrule
\end{tabular}
\caption{\label{t:results-valid} Global, Unseen and Variant MWE-based scores on validation datasets.}
\vspace*{-\baselineskip}
\end{table}

\section{Experiments}
\label{sec:exp}

We adapted the sequential labelling scheme  of \newcite{rohanian2019bridging} which is similar to IOB with the difference that it introduces a new soft label \texttt{o-} for the tokens that are in between components of an MWE. We preserved MWE categories by suffixing the label with the category name. In this case, the annotations for the idiomatic verbal expression (of type \texttt{VID}) in the sentence \textit{I would \textbf{give} this job \textbf{a go}}, would be:
$I_{\texttt{[O]}}$ $would_{\texttt{[O]}}$ 
$give_{\texttt{[B-VID]}}$ $this_{\texttt{[o-VID]}}$  $job_{\texttt{[o-VID]}}$ $a_{\texttt{[I-VID]}}$ $go_{\texttt{[I-VID]}}$ with the labels shown as subscripts in brackets.~\footnote{Embedded MWEs can be detected only if the nested MWE is not part of the nesting one and their categories are different.} 

In the development phase of the shared task, we trained various configurations of our system and evaluated the performance on development sets. Specifically, we examined the performance of our model in two settings: (1) the model is back-propagated only based on $Loss_{mwe}$ ({\em single-task}), and (2) the learning is based on the multi-task $Loss$ ({\em multi-task}). We decided on the setting to be used for each language separately based on the performance on development sets. We used \textit{bert-base-multilingual-cased} as the pre-trained model for all languages.\footnote{We tried uncased multilingual models, for {\tt FR} and {\tt PL} in particular, but we didn't observe any improvements.} Due to lack of time and resources, we did not perform any extensive hyper-parameter search. We empirically chose learning rate $3\times10^{-5}$ and batch size $10$ (except for {\tt GA} where the selected batch size is $1$). We trained the models for $10$ epochs, and the maximum lengths of sentences for training were chosen for each language separately based on the word piece tokenisation of multilingual BERT.~\footnote{When tokenisation splits words into multiple pieces, we took the prediction for the first piece as the prediction for the word. We masked the rest in the learning process.}

Table \ref{t:results-valid} shows results on the development sets. According to the shared task criteria, we report MWE-based precision, recall and F1 measures for all VMWEs and unseen ones in particular. We also consider the scores on the expressions which are syntactic variants of their occurrences in the training data useful to be reported. We chose the best setting for each language based on F1 scores on unseen VMWEs (in bold). The systems marked by * (in Table \ref{t:results-valid}) are trained after the evaluation period; therefore, their scores on test are not available in the official evaluation report.  
In the multi-task setting we tried two $\alpha$ values: $\frac{1}{300}$ and $\frac{1}{700}$. We used the value that worked best for each language ($\frac{1}{300}$ for {\tt EL}, {\tt RO}, {\tt SV} and {\tt TR}, and  $\frac{1}{700}$ for {\tt DE}, {\tt EU}, {\tt FR}, {\tt GA}, {\tt HE}, {\tt HI}, {\tt IT}, {\tt PL}, {\tt PT} and {\tt ZH}). The best model for each language was trained on both train and dev sets. The results obtained on test data are reported in Section \ref{sec:analysis}.

After the evaluation period, we also fine-tuned the dependency-CRF branch of the model on some portions of extra data 
for several lower-resource languages (e.g. {\tt GA}, {\tt HI}, {\tt HE} and {\tt TR}). We saw no notable improvement except for {\tt TR} as reported in Table~\ref{t:results-valid} (multi-taks + extra data). 
We only fine-tuned the model to learn unlabeled trees for dependency arcs, which are made available for additional data as part of the shared task. 
Due to being limited by the amount of computational power, we only partially used the extra unannotated data; therefore we leave the experiments on their effects to future work.

\section{Results and Analysis}
\label{sec:analysis}

Table \ref{t:results-test-summary} shows the summary results of our system MTLB-STRUCT on the 
test sets. For each language, we report the employed system (single or multi-task), the ratio of unseen data in the test set, global and unseen MWE-based F1 scores, and finally the system's rank ($\#$) in the open track of the shared task.\footnote{More detailed results (including precision and recall values, and token-based performance measures) are available on the shared task web page: \url{http://multiword.sourceforge.net/PHITE.php?sitesig=CONF&page=CONF_02_MWE-LEX_2020___lb__COLING__rb__&subpage=CONF_50_Shared_task_results}} Our system is applied to all $14$ languages and achieves the highest F1 score overall.

\begin{table}[t]
\centering
\small
\begin{tabular}{|l l | c | c c c || l l | c | c c c | }
\toprule
\multirow{2}{*}{Lang} & \multirow{2}{*}{System} & & {Global} & {Unseen} & &  \multirow{2}{*}{Lang} & \multirow{2}{*}{System} & & Global & Unseen &  \\
  & & unseen \% & F1 &  F1 & \# & & & unseen \% & F1 & F1 & \#  \\
\midrule
{DE} & single & 37\% & 76.17 & 49.34 & 1 & IT & single & 29\% & 63.76 & 20.81 & 3 \\ 
{EL} & multi & 31\% & 72.62 & 42.47 & 1 & PL & single & 22\% & 81.02 & 39.94 & 2 \\
EU & multi & 15\% &  80.03 & 34.41 & 1 & PT & multi & 24\% & 73.34 & 35.13 & 1 \\
FR & single & 22\% & 79.42 & 42.33 & 2 & RO & single & 7\% & 90.46 & 34.02 & 2  \\
GA & multi & 69\% & 30.07 & 19.54 & 1 & SV & multi & 31\% & 71.58 & 42.57 & 1 \\
HE & multi & 60\% & 48.30 & 19.59 & 1 & TR & multi & 26\% & 69.46 & 43.66 & 2 \\
HI & single & 45\% & 73.62 & 53.11 & 1 & ZH & multi & 38\% & 69.63 & 56.2 & 2  \\
\midrule
\multicolumn{6}{|c}{} & Overall & - & - & 70.14 & 38.53 & 1  \\

\bottomrule
\end{tabular}
\caption{\label{t:results-test-summary} The percentage of unseen expressions (unseen \%), and Global and Unseen MWE-based F1 results for all languages (Lang) in test. Column \# indicates the ranking of our system in the shared task.}
\vspace*{-\baselineskip}
\end{table}

The amount of MWEs seen in the training data is the largest contributing factor, as the percentage of seen-in-train gold MWEs is highly linearly correlated ($r=0.90$) with the global MWE-based F1 score across the languages. We achieve the highest performance in terms of MWE-based F1 score on unseen data for $8$ out of $14$ languages, with the largest gaps in performance observed on {\tt PT}, where our system outperforms {\tt Seen2Unseen} by $21.59$ points, and on {\tt HI}, where the gap between our system's F1 and that of {\tt Seen2Unseen} equals $10.45$ points. We note that our system works significantly better than the second best systems for smaller datasets ({\tt GA}, {\tt HE}, and {\tt HI}) which also happen to have larger amount of unseen expressions. At the same time, {\tt TRAVIS-mono} outperforms our system on {\tt FR}, {\tt IT}, {\tt PL}, {\tt TR}, and {\tt ZH}, with the largest gap of $5.68$ points observed on {\tt FR}. 

In addition, our system's performance is balanced across {\em continuous} and {\em discontinuous MWEs}, with the exceptions of {\tt HI} and {\tt TR}, where discontinuous MWEs amount to $7\%$ and $4\%$ of all MWEs, respectively, and our system's performance drops by as much as $30$ F1 points compared to its performance on continuous MWEs. The distinction between {\em multi-} and {\em single-token MWEs} is only applicable to $3$ languages, on two of which ({\tt DE} and {\tt SV}) our system achieves an F1 score above $0.80$ on {\em single tokens}.

Finally, the shared task data shows a wide diversity of VMWE categories present in different languages: from just three in {\tt EU} and {\tt TR} up to eight in {\tt IT}. Once again, we note that our system is applicable to detection of all categories: for instance, it achieves the highest F1 scores among all systems in identification of {\tt LS.ICV}, a rare language-specific category of inherently clitic verbs used only in Italian. At the same time, we identify {\tt LVC.cause}, light-verb constructions with the verb adding a causative meaning to the noun, as the most problematic category on which our system achieves comparatively poorer results, especially on {\tt DE}, {\tt EL}, {\tt FR}, {\tt HI}, {\tt PT}, and {\tt SV}.

It is worth noting that no language specific feature is used in our system and the authors were not involved in the creation of the datatsets.
Overall, we note that our system is not only cross-lingual, but also robust in terms of its performance and is capable of generalising to unseen MWEs. 
\section{Conclusions and Future Work}
We described MTLB-STRUCT, a semi-supervised system that is based on pre-trained BERT masked language modelling and that jointly learns VMWE tags and dependency parse trees. The system ranked first in the open track of the PARSEME shared task - edition 1.2 and shows the overall state-of-the-art performance for detecting unseen VMWEs. 
In future, we plan to augment the dependency parsing architecture to train on dependency relation categories (labels) as well as dependency arcs. We also plan to improve our system by making it more efficient in order to train the dependency parsing module on the extra available unannotated datasets.

\section*{Acknowledgments}

This paper reports on research supported by Cambridge Assessment, University of Cambridge. We are grateful to the anonymous reviewers for their valuable feedback. We gratefully acknowledge the support of NVIDIA Corporation with the donation of the Titan V GPU used in this research. 
 
\bibliographystyle{coling}
\bibliography{coling2020}

\end{document}